\documentclass[conference]{IEEEtran}

\usepackage[pdftex]{graphicx}
\usepackage[dvipsnames]{xcolor}
\usepackage[hidelinks]{hyperref}
\definecolor{linkcolor}{HTML}{648EB0}
\newcommand{\link}[2]{\href{#1}{\textcolor{linkcolor}{#2}}}
\begin{document}

%+++++++++++++++++++++++++++++++++++++++++++++++++++
%\title{\Large o80: Bridging Robotics C++ Realtime Control to Machine Learning Python }

\title{\Large Synchronizing Machine Learning Algorithms, Realtime Robotic Control and Simulated Environment with o80}

\author{
Vincent~Berenz, Felix~Widmaier, Simon~Guist, Bernhard~Schölkopf and Dieter~Büchler \\
Max Planck Institute for Intelligent Systems, Tübingen, Germany \\
\\
\includegraphics[width=13cm,keepaspectratio]{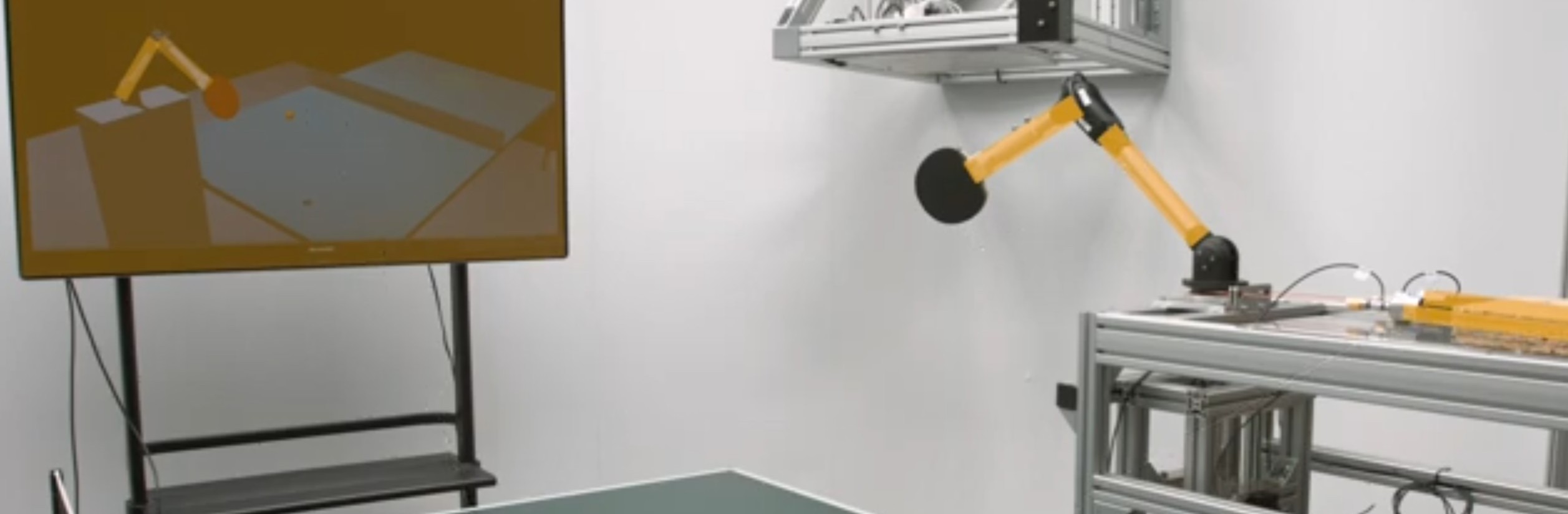}
\\
\\
(work presented at the Robot Software Architectures Workshop -  \link{https://roboticsa.github.io/RoboticSA2023/}{RSA 2023}, ICRA)
}

%+++++++++++++++++++++++++++++++++++++++++++++++++++

\maketitle

% ================
% # Abstract     #
% ================

\begin{abstract}
Robotic applications require the integration of various modalities, encompassing perception, control of real robots and possibly the control of simulated environments. While the state-of-the-art robotic software solutions such as ROS 2 provide most of the required features, flexible synchronization between algorithms, data streams and control loops can be tedious. 
o80 is a versatile C++ framework for robotics which provides a shared memory model and a command framework for real-time critical systems. It enables expert users to set up complex robotic systems and generate Python bindings for scientists. o80's unique feature is its flexible synchronization between processes, including the traditional blocking commands and the novel ``bursting mode'', which allows user code to control the execution of the lower process control loop. This makes it particularly useful for setups that mix real and simulated environments.

\end{abstract}

% ========================
% # I. Introduction      #
% ========================

\section{Introduction}

o80\footnote{`o80' is a monogram representing two running processes (`o' and `0') and the communication between them (`8')} is an open-source C++ toolbox that allows expert users to create custom Python API suitable for interacting with complex robotic setup \cite{Berenz2021}. o80 provides functions for:

\begin{itemize}
    \item the spawning realtime processes (e.g., running on RT-Preempt, low latency kernel or Xenomai)
    \item the synchronization of these processes 
    \item the asynchronous access to a shared memory hosting the history of all sensor data
    \item the sending of custom commands, including blocking and not blocking commands (see section \ref{commands})
    \item the automated generation of customized Python bindings
\end{itemize}

While o80 shares similarities with ROS (spawning of processes and C++/Python interoperability) and actionlib (management of commands) \cite{ros}\cite{actionlib}, it has differences to them: it relies on a shared memory model rather than a publish-subscribe model. But the core difference, and novelty of o80, is its flexibility regarding synchronization. When using o80, users may either synchronize their higher level code with the lower level control process via blocking commands (see section \ref{commands}). Alternatively, it is possible to synchronize the lower level control process to the higher level code via the new `bursting mode'. In bursting mode, the low-level control process blocks until the user process sends a request to run one or more iterations. This unique feature is useful when interacting with a simulated robot or even, as shown in section \ref{HYSR}, an experimental setup involving a real robot and a simulated environment. 

However, as opposed to ROS, o80 does not support network communication, as it requires the processes it orchestrates to run on the same computer.

The o80 framework is a two-levels system. The first level involves the expert user, who is responsible for implementing the C++ driver classes that are specific to the hardware used in the experiment. During compilation, o80 utilizes these classes as templates to generate executables that create the real-time control processes. 
In addition to implementing the driver classes, the expert user is responsible for generating a Python API tailored to the needs of the users. o80 allows for automated generation of Python bindings. As a result, the users will have access to a simple Python interface. These users can focus on designing experiments without being burdened by the implementation details of the robotic setup.

\section{Overview}

\subsection{Backend, frontend and shared memory}

o80 is based on the interaction between:

\begin{itemize}
    \item a backend, i.e., a process which communicates in realtime with a hardware device. It is responsible for sending commands to the device and receiving observations from it. Upon spawning, a backend creates a dedicated shared memory. It uses this shared memory to 1) read user commands and 2) write sensor and/or state related data.
    \item a frontend provides methods for connecting to a related backend's shared memory for 1) writing user commands and 2) for reading data. 
\end{itemize}

%The end-users of an o80 based system (i.e. scientists) are expected to interact with the setup via Python instances of frontends.

\subsection{Realtime control and commands}\label{commands}

Stable control on realtime critical systems requires high-frequency software loops.
o80 backend processes are developed in C++ to comply with real-time constraints, calculating the desired state for each actuator at each iteration.
Frontends provide Python methods that send higher-level, non-realtime commands to the backend's shared memory. These commands specify implicit desired state trajectories that rely on interpolation (based on specified durations, speeds, or numbers of server iterations). For instance, if a user command specifies that a robot should reach a desired state over a specified duration, the backend will generate the higher-frequency low-level commands that interpolate from the robot's current state to the desired state. By translating the frontend's commands into low-level commands that operate the system at the required frequency, the backend ensures stable control of the real-time critical systems. The frontend's API is flexible and allows for queuing or interrupting commands, as well as issuing both blocking and non-blocking commands.

\subsection{Reading observation}\label{observation}
The backend writes current actuator state and custom sensor information to shared memory at each control iteration, which can be retrieved using various methods provided by the frontend API. Users can request the latest information, information from past server iterations, or wait for future data using a blocking method (which can be used for synchronizing user processes with backend, see section \ref{bursting}). Multiple instances of the frontend can read asynchronously from shared memory. For example, this enables users to run logging scripts for the robot's state in parallel with control scripts.

\subsection{Embedding backends}
In addition to generating executables that spawn backend processes, o80's API also supports embedding instances of C++ or Python backends in other processes. This feature can be utilized to extend o80's functionality to simulations. Section \ref{HYSR} provides an example of o80 backends being used to control the movement of bodies in a Mujoco environment.

\subsection{Synchronization and bursting mode}\label{bursting}
o80 provides two synchronization modes: "normal" and "bursting". In normal mode, the backend process runs in real-time at its required frequency, while the user Python process can synchronize with it through the blocking waiting methods mentioned earlier. However, in bursting mode, the backend process blocks until the frontend requires it to run one or more iterations. Bursting mode is typically used when interacting with a simulator. The frontend allows users to create commands that require several backend iterations to execute, which can then be executed as fast as the simulator allows.

\section{HYSR training of table tennis playing robot}\label{HYSR}

A team of researchers from the Max Planck Institute for Intelligent Systems is exploring the potential of reinforcement learning for teaching a robotic arm, actuated by a pneumatic artificial muscle (PAM), to play table tennis~\cite{learningscratch}. The scientists are using a hybrid simulation and real training (HYSR) technique that involves mirroring the movements of a real robotic arm with a Mujoco simulated arm. This approach allows the real robot to interact with a single or multiple simulated balls that are being replayed from recorded ball trajectories, facilitating practical long-term learning of table tennis~\footnote{\link{https://youtu.be/GQtpSMEpn5A}{Video} of HYSR training.}. 
Additionally, virtual environments can be adapted for data-efficient training by, for instance, playing with multiple virtual balls~\cite{guist_hindsight_2023}.

To set up the experiment, the researchers required:
\begin{itemize}
    \item A real-time control process that sends pressure commands to the PAM controllers of the real robot at a fixed frequency of 500Hz.
    \item A Mujoco simulated robot that mirrors the movements of the real robot and replays recorded ball trajectories. Each iteration of the Mujoco simulator takes 0.02 seconds.
    \item A GYM reinforcement learning environment with a step function running at 100Hz.
    \item Control of other hardware for real ball experiments, including a Vicon system for localizing the robot and table, a ball launcher, and an RGB-based ball detection system.
\end{itemize}
  
o80 allowed to solve all the synchronization issues related to this setup. A backend process runs at 500Hz and controls the real pneumatic muscles while reading related robot states. A backend instance, running in bursting mode, is embedded in the Mujoco simulated environment. Frontends, connected to both backends, are embedded in the learning environment, asynchronously sending pressure actions and reading states to/from the real robot, sending mirroring states to the simulated robot, and sending bursting commands to the Mujoco simulated environment.

In addition, o80 simplified the process of spawning new processes that create additional frontends, which can easily access the shared memory history to log data, visualize the robot state in real-time, and monitor both the simulated and real robot state.

The code and documentation of this project are available as open source online \cite{pamsource}.

\section{Conclusion}
o80 is a versatile middleware system that offers flexible control of robotic systems. It allows expert users to develop a user-friendly Python API that makes it easier for machine learning scientists to use complex robotic setups.  Its shared memory model, different synchronization modes, and interpolation-based command framework distinguish it from ROS.
For more information and code examples, we refer to o80's comprehensive documentation \cite{o80source}. 

\bibliographystyle{IEEEtran}
\bibliography{IEEEabrv,biblio}

\begin{thebibliography}{1}
\providecommand{\url}[1]{#1}
\csname url@samestyle\endcsname
\providecommand{\newblock}{\relax}
\providecommand{\bibinfo}[2]{#2}
\providecommand{\BIBentrySTDinterwordspacing}{\spaceskip=0pt\relax}
\providecommand{\BIBentryALTinterwordstretchfactor}{4}
\providecommand{\BIBentryALTinterwordspacing}{\spaceskip=\fontdimen2\font plus
\BIBentryALTinterwordstretchfactor\fontdimen3\font minus
  \fontdimen4\font\relax}
\providecommand{\BIBforeignlanguage}[2]{{%
\expandafter\ifx\csname l@#1\endcsname\relax
\typeout{** WARNING: IEEEtran.bst: No hyphenation pattern has been}%
\typeout{** loaded for the language `#1'. Using the pattern for}%
\typeout{** the default language instead.}%
\else
\language=\csname l@#1\endcsname
\fi
#2}}
\providecommand{\BIBdecl}{\relax}
\BIBdecl

\bibitem{Berenz2021}
\BIBentryALTinterwordspacing
V.~Berenz, M.~Naveau, F.~Widmaier, M.~Wüthrich, J.-C. Passy, S.~Guist, and
  D.~Büchler, ``The o80 c++ templated toolbox: Designing customized python
  apis for synchronizing realtime processes,'' \emph{Journal of Open Source
  Software}, vol.~6, no.~66, p. 2752, 2021. [Online]. Available:
  \url{https://doi.org/10.21105/joss.02752}
\BIBentrySTDinterwordspacing

\bibitem{ros}
\BIBentryALTinterwordspacing
{Stanford Artificial Intelligence Laboratory et al.}, ``Robotic operating
  system.'' [Online]. Available: \url{https://www.ros.org}
\BIBentrySTDinterwordspacing

\bibitem{actionlib}
\BIBentryALTinterwordspacing
M.~Carroll, J.~Perron, E.~Marder-Eppstein, V.~Pradeep, and M.~Arguedas,
  ``{actionlib},'' 2009. [Online]. Available:
  \url{http://wiki.ros.org/actionlib}
\BIBentrySTDinterwordspacing

\bibitem{learningscratch}
D.~B{\"u}chler, S.~Guist, R.~Calandra, V.~Berenz, B.~Sch{\"o}lkopf, and
  J.~Peters, ``Learning to play table tennis from scratch using muscular
  robots,'' \emph{IEEE Transactions on Robotics (T-RO)}, vol.~38, no.~6, pp.
  3850--3860, 2022.

\bibitem{guist_hindsight_2023}
S.~Guist, J.~Schneider, A.~Dittrich, V.~Berenz, B.~Sch{\"o}lkopf, and
  D.~B{\"u}chler, ``Hindsight states: Blending sim and real task elements for
  efficient reinforcement learning,'' \emph{arXiv preprint arXiv:2303.02234},
  2023.

\bibitem{pamsource}
\BIBentryALTinterwordspacing
V.~Berenz, F.~Widmaier, S.~Guist, and D.~Büchler, ``{PAM} robot software
  documentation,'' 2020. [Online]. Available:
  \url{https://intelligent-soft-robots.github.io/pam\_documentation/}
\BIBentrySTDinterwordspacing

\bibitem{o80source}
\BIBentryALTinterwordspacing
V.~Berenz, S.~Guist, and D.~Büchler, ``{o80} robot software documentation,''
  2020. [Online]. Available:
  \url{http://people.tuebingen.mpg.de/mpi-is-software/o80/docs/o80/index.html}
\BIBentrySTDinterwordspacing

\end{thebibliography}

\end{document}